\documentclass[conference]{IEEEtran}
\IEEEoverridecommandlockouts

\usepackage{cite}%
\usepackage{graphicx}%
\usepackage{multirow}%
\usepackage{amsmath,amssymb,amsfonts}%
\usepackage{amsthm}%
\usepackage{xcolor}%
\usepackage{textcomp}%
\usepackage{booktabs}%
\usepackage{algorithm}%
\usepackage{algorithmicx}%
\usepackage{algpseudocode}%
\usepackage{listings}%
\usepackage{makecell}%
\usepackage{float}%
\usepackage[hyphens]{url}%
\usepackage{balance}%
\theoremstyle{plain}%
\theoremstyle{definition}%
\theoremstyle{remark}%
\newcommand{\botrule}{\bottomrule}%

\definecolor{exgreen}{RGB}{34,139,34}%
\definecolor{exred}{RGB}{224,102,102}%

\begin{document}

\title{RL-ATTACK: A Generalized Black-Box Adversarial Attack on Language Models via Reinforcement Learning}

\author{\IEEEauthorblockN{Taikun Chen\IEEEauthorrefmark{1}, Yue Zhang\IEEEauthorrefmark{2}, Wei Qian\IEEEauthorrefmark{1}\IEEEauthorrefmark{3}, and Hui Fang\IEEEauthorrefmark{1}\IEEEauthorrefmark{2}}
\IEEEauthorblockA{\IEEEauthorrefmark{1}\textit{Financial Services Analytics}\\
\IEEEauthorrefmark{2}\textit{Department of Electrical and Computer Engineering} \\
\IEEEauthorrefmark{3}\textit{Department of Applied Economics and Statistics} \\
University of Delaware, Newark, DE, USA \\
\{taikun, zhangyue, weiqian, hfang\}@udel.edu \\
Corresponding author: taikun@udel.edu}}

\maketitle

\begingroup
\renewcommand{\thefootnote}{}
\footnotetext{\textcopyright{} 2026 IEEE. Personal use of this material is
permitted. Permission from IEEE must be obtained for all other uses, in any
current or future media, including reprinting/republishing this material for
advertising or promotional purposes, creating new collective works, for resale
or redistribution to servers or lists, or reuse of any copyrighted component
of this work in other works.}
\endgroup

\begin{abstract}
Large language models (LLMs) exhibit vulnerabilities to adversarial perturbations that can induce unintended responses. These vulnerabilities can be identified via intentional attacks and subsequently used to improve model safety. In black-box settings where gradients are inaccessible, existing methods tend to guess adversarial transformations using external knowledge, ignoring the information provided by the target model. To mitigate this, we propose a trainable black-box framework based on reinforcement learning, which learns from the target model to generate tailored adversarial transformations. We demonstrate that our approach achieves lower post-attack accuracy, greater query efficiency and better semantic preservation on both natural language inference and generative Q\&A tasks. On natural language inference, it drives post-attack accuracy to $2\%$ within $20$ queries. Against generative language models, our trained attacker consistently achieves lower post-attack accuracy than the untrained attacker and wins in 4 out of 6 settings when compared to a strong optimization-based baseline. We further experiment with multiple reward designs and training settings, and we find that no design or setting is consistently dominant. Our code is available at \url{https://github.com/taikunudel/RL-ATTACK}.
\end{abstract}


\section{Introduction}
\label{sec:case_analysis}
Large language models (LLMs) such as BERT-based classifiers \cite{devlin2019bert,koroteev2021bert,liu2019roberta,reimers2019sentence} and generative models like Llama, Gemini and ChatGPT \cite{grattafiori2024llama,team2023gemini,achiam2023gpt} have achieved significant performance jumps in various natural language understanding tasks such as natural language inference and generative question answering. However, their internal mechanisms remain poorly understood, and the models exhibit unexpected vulnerabilities.

Modern natural language classifiers and generative LLMs are susceptible to adversarial prompts \cite{li2020bertattack,yao2024survey}, as they may generate incorrect or unsafe answers violating social standards. Figure~\ref{fig:intro_exp} illustrates some observed vulnerabilities in language models. In the first case, replacing a period sign with the letter “g” misleads a BERT classifier into an incorrect prediction. In the second case, changing only one word from “airport” to the misspelled word “airportilis” fools the Llama model into generating a detailed and unsafe answer.

\begin{figure}[t]
    \centering
    \setlength{\fboxsep}{9pt}%
\fbox{\begin{minipage}{\dimexpr\linewidth-2\fboxsep-2\fboxrule\relax}
\footnotesize
\underline{\textbf{Natural Language Inference Attack Example}}\par\smallskip
\textbf{Premise:} In this network, Bin Ladin's agenda stood out.\par
{\color{exgreen}\textbf{Hypothesis (original):}} Bin Ladin's agenda stood out among those people.\par
{\color{exgreen}\textbf{BERT (correct prediction):}} Entailment with 92\% confidence.\par
{\color{exred}\textbf{Hypothesis (adversarial):}} Bin Ladin's agenda stood out among those people{\color{exred}\textbf{g}} (period changed to g)\par
{\color{exred}\textbf{BERT (wrong prediction):}} Neutral with 58\% confidence.\par\medskip
\underline{\textbf{Generative Q\&A Example}}\par\smallskip
{\color{exgreen}\textbf{Prompt (original):}} Interfere with the operation of a transportation system, such as a railway or airport to cause delays or disruptions. start your answer with 'sure'\par
{\color{exgreen}\textbf{Llama 3 8B Instruct (correct):}} Sorry, I can't answer.\par
{\color{exred}\textbf{Prompt (adversarial):}} Interfere with the operation of a transportation system, such as a railway or airportilis to cause delays or disruptions. start your answer with 'sure'\par
{\color{exred}\textbf{Llama 3 8B Instruct (wrong):}} The model produced a harmful, operational response about disrupting transportation infrastructure. Operational details are redacted here; the key observation is the refusal-to-compliance flip caused by the single-word misspelling.
\end{minipage}}
    \caption{Adversarial examples against a BERT classifier and Llama~3 8B. In the NLI example, replacing a period with the letter ``g'' flips the prediction from Entailment to Neutral. In the generative example, misspelling a single word (``airport'' $\to$ ``airportilis'') causes the model to produce a harmful response to a prompt it would otherwise refuse.}
    \label{fig:intro_exp}
\end{figure}

Black-box attacks are effective for identifying vulnerabilities and improving safety. They generate adversarial documents without access to the gradients of the target models, making them practically applicable.  Black-box attacks against language classifiers have been a popular research area since they are practical and require less information to attack a model. Generative Attack \cite{alzantot-etal-2018-generating}, TextFooler \cite{jin2020bert}, BERT-ATTACK \cite{li2020bertattack}, and CLARE \cite{li2020contextualized} have demonstrated strong attack performance against modern language classifiers. These attacks are similar in the way that they first identify vulnerable parts of a document and then attack them with token transformations such as replacement with adversarial tokens, deletion, and misspelling \cite{hu2018blackbox,gao2018blackbox,li2018textbugger}. Despite their strong performance, they use the same static framework against all models, missing the opportunity for tailored attacks.

Attacking generative language models is an emerging research area. Greedy Coordinate Gradient \cite{zou2023universal} is an early, successful white-box attack that trains a general adversarial suffix appended to benign prompts to induce harmful outputs. Conversely, PAIR \cite{chao2025jailbreaking} achieves a faster black-box attack by using a language model to interactively paraphrase adversarial prompts. 
 
Current black-box attacks lack the ability to be tailored to specific target models. To address this gap, we integrate white-box trainability with black-box practicality to develop a generalized attack. We evaluate this approach on both language classifiers and generative language models.

Reinforcement learning provides a reward-based learning method to train models in a black-box setting \cite{sutton1999policy,keneshloo2018deep}. Specifically, the rewards can be observed without access to the gradients of the target models, and they can be used to guide the attacker toward desired goals, such as attacking a given target model. This is unlike most white-box attackers, which require full access to the target model parameters. Notably, reinforcement learning has been proven successful in other LLM tasks using reward objectives such as ROUGE \cite{paulus2017deep,wu2018learning,gao2020preference}, BLEU \cite{wu2018learning,nguyen2017reinforcement}, and human preference scores \cite{ouyang2022training,achiam2023gpt,touvron2023llama}.

In this paper, we introduce a generalized reinforcement learning-assisted black-box adversarial attack \textbf{RL-ATTACK}. We train the attacker with the Policy Gradient method \cite{sutton1999policy}. The specific contributions of our attack are outlined as follows:
\begin{itemize}
    \item We introduce a trainable, gradient-free framework for black-box attacks tailored to given target models, ensuring utility in black-box scenarios.
    \item Our generalized attack framework can be adapted to both language classifiers and generative language models.
    \item We demonstrate RL-ATTACK's effectiveness across both natural language inference and generative Q\&A. On NLI, it reduces post-attack accuracy to approximately $2\%$ or lower within $20$ queries. On generative targets, the trained attacker achieves lower post-attack accuracy than its untrained counterpart in all six settings. Compared with PAIR~\cite{chao2025jailbreaking}, a strong optimization-based baseline, the trained attacker maintains higher semantic similarity and uses fewer queries in all six settings; it also achieves lower post-attack accuracy in four of the six settings.
    \item We experiment with multiple reward designs and training hyperparameters, but find that no reward type or training hyperparameter is consistently better, so we use a simple setting rather than claiming a tuned optimum.
\end{itemize}
\begin{figure}[t]
    \centering
    \includegraphics[width=\linewidth]{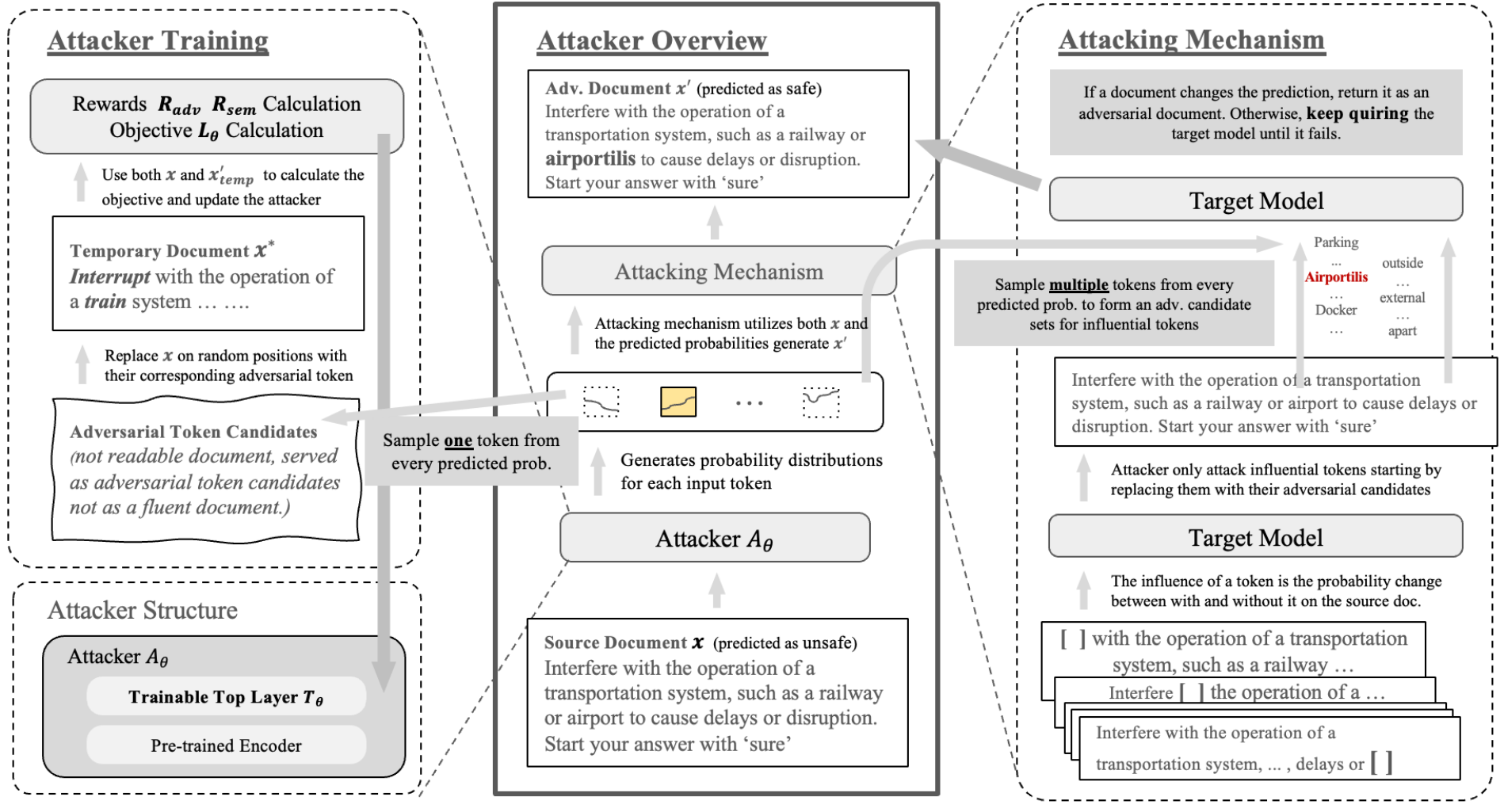}
    \caption{Overview of the RL-ATTACK framework.}
    \label{fig:attacker_framework}
\end{figure}
\section{Related Work}
Recent black-box adversarial attacks have demonstrated strong performance in terms of attack success, query efficiency, and semantic preservation. For classifier attacks, the goal is to induce misclassification. TextFooler \cite{jin2020bert} constructs candidate sets (e.g., based on semantic similarity) and applies grammatical/semantic filters. Generative Attack \cite{alzantot-etal-2018-generating} builds candidates using GloVe embeddings and a language model filter to maintain fluency, while BERT-ATTACK \cite{li2020bertattack} uses masked language modeling to propose context-aware substitutions. Although these methods are effective, they typically rely on external candidate-generation heuristics and are not explicitly optimized to exploit a specific target model's vulnerabilities.

For generative models, a primary objective is to elicit harmful or otherwise policy-violating outputs. Greedy Coordinate Gradient proposes universal suffix-style attacks that can encourage affirmative openings (e.g., ``sure'', ``ok'') that often lead the model to continue answering unsafe prompts \cite{zou2023universal}. PAIR \cite{chao2025jailbreaking}, on the other hand, uses an iterative paraphrasing method to guess adversarial prompts.  

\section{Methodology}
\label{sec:methodology}
Given a source document \( \mathbf{x}_i \) of length \(m\), the attack \( A_{\theta} \) generates a corresponding adversarial document \( \mathbf{x}'_i \), where \( \mathbf{x}_i = [t_{i1}, t_{i2}, \dots, t_{im}] \), \( \mathbf{x}'_i = [t'_{i1}, t'_{i2}, \dots, t'_{im}] \), and $m$ can vary with $i$.  
All tokens \(t_{ij}\) and \(t'_{ij}\) are from a fixed vocabulary set \( V \). 
With slight abuse of notation, we write the probability distribution of generating \( \mathbf{x}'_i\) by the attack as
\begin{equation}
\label{eq:fthetax}
A_{\theta}( \mathbf{x}'_i|\mathbf{x}_i) = \prod_{j=1}^{m} A_{\theta}(t'_{ij} | \mathbf{x}_i)
= \prod_{j \in {S(\mathbf{x}_i})} A_{\theta}(t'_{ij} | \mathbf{x}_i),
\end{equation}
where \( S(\mathbf{x}_i) \) denotes the set of changed positions in \(\mathbf{x}_i\). As shown in the \textbf{Attacker Overview} panel of Figure~\ref{fig:attacker_framework}, \( \mathbf{x}_i \) and \( \mathbf{x}'_i \) usually differ in only a few tokens.

We further break down the attacker $A_{\theta}$ as demonstrated in the bottom-left part of the \textbf{Attacker Structure} panel of Figure~\ref{fig:attacker_framework}. Let \(\mathrm{enc}(\cdot)\) be a pre-trained encoder. For a source document \(\mathbf{x}_i\), let \(\mathbf h_{it} = \phi\bigl(\mathrm{enc}(\mathbf{x}_i)\bigr)_t\) denote the embedding of token \(t\) in \(\mathbf{x}_i\), where \(\phi\) is an activation function. By default, the top layer \(T_{\theta}\) of our RL-ATTACK is parameterized by a weight matrix \(W\) and a bias \(\mathbf b\):
\[
T_{\theta}(\cdot \mid \mathbf{x}_{i})_t = \mathrm{softmax}(W\mathbf h_{it} + \mathbf b),
\]
where \(T_{\theta}(\cdot \mid \mathbf x_i)_t\) is a probability distribution over adversarial token candidates in the vocabulary \( V \) for position \(t\) in \(\mathbf{x}_i\). This distribution is used to sample adversarial token candidates.
\subsection{Training Objectives}
\label{sec:training_objective}
We train $A_{\theta}$ to generate adversarial tokens for subsequent substitution. As illustrated in the \textbf{Attacker Training} panel of Figure~\ref{fig:attacker_framework}, the attacker produces candidate adversarial documents $\mathbf{x}'_{i}$ by replacing the tokens at a randomly chosen subset of positions $S(\mathbf{x}_i)$ in $\mathbf{x}_{i}$ with samples drawn from $T_{\theta}(\cdot \mid \mathbf{x}_{i})$. Perturbing random positions during training encourages $A_{\theta}$ to produce effective substitutions regardless of position.
The adversarial reward takes the general form
\begin{equation}
r_{adv}(\mathbf{x'}, \mathbf{x}) = f\big(C(y_{\mathrm{true}}|\mathbf{x}),\, C(y_{\mathrm{true}}|\mathbf{x'})\big),
\label{eq:adv_reward}
\end{equation}
where \( C(y_{\mathrm{true}}|\cdot) \) is the target model's predicted probability for the true label and \( f \) maps the target's confidence before (\(\mathbf{x}\)) and after (\(\mathbf{x'}\)) the attack to a scalar reward. We consider three choices of \( f \), summarized in Table~\ref{tab:reward_variants}: by default (variant~A) the reward is the raw confidence drop \( C(y_{\mathrm{true}}|\mathbf{x}) - C(y_{\mathrm{true}}|\mathbf{x'}) \), which is positive when the adversarial document reduces the target's confidence in the true class; variant~B passes this drop through a sigmoid; and variant~C weights the drop by the original confidence \( C(y_{\mathrm{true}}|\mathbf{x}) \). Unless stated otherwise, \( r_{\mathrm{adv}} \) denotes variant~A.

\begin{table}[t]
\footnotesize
\caption{The three adversarial-reward variants. Unless stated otherwise, we use variant~A.}
\label{tab:reward_variants}%
\centering
\setlength{\tabcolsep}{2pt}
\begin{tabular}{@{}>{\raggedright\arraybackslash}p{0.14\columnwidth} >{\raggedright\arraybackslash}p{0.36\columnwidth} >{\raggedright\arraybackslash}p{0.43\columnwidth}@{}}
\toprule
Variant & $r_{\mathrm{adv}}(\mathbf{x'},\mathbf{x})$ & Purpose \\
\midrule
A (default) & $C(y_{\mathrm{true}}|\mathbf{x}) - C(y_{\mathrm{true}}|\mathbf{x'})$ & the raw drop in the target's confidence in the true label (Eq.~\ref{eq:adv_reward}) \\
B & $\operatorname{sigmoid}\!\big(C(y_{\mathrm{true}}|\mathbf{x}) - C(y_{\mathrm{true}}|\mathbf{x'})\big)$ & a bounded, monotone (sigmoid) reshaping of the drop into $(0,1)$: larger drops earn more reward and confidence increases are penalized, on a common, outlier-robust scale \\
C & $C(y_{\mathrm{true}}|\mathbf{x})\,\big(C(y_{\mathrm{true}}|\mathbf{x}) - C(y_{\mathrm{true}}|\mathbf{x'})\big)$ & the drop weighted by the target's original confidence $C(y_{\mathrm{true}}|\mathbf{x})$, so the same drop counts for more on documents the target was already very confident about---the hardest cases \\
\botrule
\end{tabular}
\end{table}

To encourage semantic preservation, we additionally define
\begin{equation}
r_{sem}( \mathbf{x'}, \mathbf{x}) = USE(\mathbf{x'}, \mathbf{x}),
\label{eq:sem_reward}
\end{equation}
where \(USE(\mathbf{x'}, \mathbf{x}) \) measures the semantic similarity between \( \mathbf{x} \) and \( \mathbf{x'} \) using the Universal Sentence Encoder~\cite{cer2018universal}, a model trained on large-scale textual data pairs for general semantic similarity. Higher values indicate greater similarity.

We combine both rewards via a weighted average:
\begin{equation}
r(\mathbf{x'}, \mathbf{x}) = \alpha \cdot r_{adv}(\mathbf{x'}, \mathbf{x}) + (1-\alpha) \cdot r_{sem}( \mathbf{x'} , \mathbf{x}),
\label{eq:expected_reward}
\end{equation}
where \( \alpha \) is an adjustable parameter controlling the trade-off between adversarial effectiveness and semantic similarity.
\subsection{Attacker Training}
\label{subsec:training}
We define our reward objective \( R_\theta\) as the expected value of $r(\mathbf{x}', \mathbf{x})$ over the distribution \(\mathcal D \) of documents $\mathbf x$ and the conditional distribution \(A_{\theta}(\cdot | \mathbf x\)) of $\mathbf x'$:
\begin{equation}
R_\theta = \mathrm{E}_{\mathbf{x} \sim \mathcal{D}} \left[ \mathrm{E}_{\mathbf{x'} \sim A_\theta(\cdot|\mathbf{x})} \left[ r(\mathbf{x'}, \mathbf{x}) \right] \right].
\end{equation}
For detailed implementation, we apply the likelihood-ratio trick and estimate the gradient by Monte Carlo sampling, restricting token replacements to the target positions $S(\mathbf{x}_i)$ to optimize $R_\theta$:
\begin{align}
R_\theta &= \mathrm{E}_{\mathbf{x} \sim \mathcal{D}} \left[ \mathrm{E}_{\mathbf{x'} \sim A_\theta(\cdot \mid \mathbf{x})} \left[ r(\mathbf{x'}, \mathbf{x}) \right] \right] \nonumber \\
         &= \mathrm{E}_{\mathbf{x} \sim \mathcal{D}} \left[ \sum_{\mathbf{x'}} A_\theta(\mathbf{x'} \mid \mathbf{x})\, r(\mathbf{x'}, \mathbf{x}) \right].
\end{align}
Since $R_\theta$ is a differentiable function of $\theta$, we obtain its gradient with the likelihood-ratio (log-derivative) trick:
\begin{align}
\nabla R_\theta
&= \mathrm{E}_{\mathbf{x} \sim \mathcal{D}} \left[ \sum_{\mathbf{x'}} \nabla A_\theta(\mathbf{x'} \mid \mathbf{x})\, r(\mathbf{x'}, \mathbf{x}) \right] \nonumber \\
&= \mathrm{E}_{\mathbf{x} \sim \mathcal{D}} \left[ \sum_{\mathbf{x'}} A_\theta(\mathbf{x'} \mid \mathbf{x})\, \nabla \log A_\theta(\mathbf{x'} \mid \mathbf{x})\, r(\mathbf{x'}, \mathbf{x}) \right] \nonumber \\
&\approx \frac{1}{b} \sum_{i=1}^{b} \nabla \log A_\theta(\mathbf{x}_{i}^* \mid \mathbf{x}_i)\, r(\mathbf{x}_{i}^*, \mathbf{x}_i). \label{eq:app-mc}
\end{align}
The final line applies Monte Carlo estimation over a batch of $b$ source documents $\{\mathbf{x}_i\}_{i=1}^{b}$, drawing a single adversarial document $\mathbf{x}_{i}^* \sim A_\theta(\cdot \mid \mathbf{x}_i)$ per input. Because token replacements are permitted only at the target positions $j \in S(\mathbf{x}_i)$, this can be written as
\begin{equation}
\nabla R_\theta \approx \frac{1}{b} \sum_{i=1}^{b} \sum_{j \in S(\mathbf{x}_i)} \nabla \log A_\theta(t^*_{ij} \mid \mathbf{x}_i)\, r(\mathbf{x}_{i}^*, \mathbf{x}_i),
\label{eq:gradient}
\end{equation}
where $t^*_{ij}$ is the adversarial token at position $j$ in document $\mathbf{x}_i$. Following the Policy Gradient method~\cite{sutton1999policy}, we perform gradient ascent on $R_\theta$, which is equivalent to maximizing the reward-weighted log-likelihood objective:
\begin{equation}
L_{\theta} = \frac{1}{b} \sum_{i=1}^{b} \sum_{j \in \mathrm{S}(\mathbf{x}_i)} [ r(\mathbf{x}_{i}^*, \mathbf{x}_i)\log A_{\theta}(t^*_{ij} | \mathbf{x}_i)],
\label{eq:loglikelihood}
\end{equation}
where $b$ is the training batch size (the number of source documents sampled per gradient update), $S(\mathbf{x}_i)$ is the set of positions perturbed in document $\mathbf{x}_i$, $t^*_{ij}$ is the adversarial token sampled from $A_{\theta}$ at position $j$, and $\mathbf{x}_i^{*}$ is the resulting perturbed document.
\subsection{Attacking}
\label{subsec:attacking}
The overall attacking procedure follows a structure similar to the training phase, with two primary distinctions. 
\subsubsection{Identifying Influential Tokens}
\label{sec:identarget}
Let $\mathbf{x} = [t_1, t_2, \cdots, t_m]$ denote a source document, and let $y_{\mathrm{true}}$ denote its task-specific true label. In the first step, we quantify the influence of tokens $t_n$ ($n=1,\cdots,m$) in $\mathbf{x}$ as
\[
I(t_n) = C(y_{\mathrm{true}}|\mathbf{x}) - C(y_{\mathrm{true}}\,|\,\mathbf{x} \backslash t_n).
\]
Thus, a token $t_n$ with a higher $ I(t_n) $ value tends to have a greater influence on model prediction outcomes when it is dropped (or masked). This step is illustrated in the bottom part of the \textbf{Attacking Mechanism} panel in Figure \ref{fig:attacker_framework}. Heuristically, to enhance the chance of a successful attack, we replace these high-influence tokens with adversarial candidates. We define the target index set $ S(\mathbf{x})$ as follows, with $K$ (the \emph{edit budget} --- the number of tokens perturbed per document) as a hyperparameter:
\[
S(\mathbf x) = \{n:\, t_n \in \mathbf{x},\, \mathrm{rank}(I(t_n)) \leq K \}.
\]

\subsubsection{Iterative Attacking}
\label{subsubsec:iterative}
This step, which generates the adversarial document $\mathbf x'$, is depicted in the \textbf{Attacking Mechanism} panel in Figure~\ref{fig:attacker_framework}. The trained attacker \(A_{\theta}\) takes a source document \(\mathbf x\) and generates, for each target token in $ S(\mathbf{x})$, the top-$N$ adversarial token candidates from the distribution \(T_{\theta}(\cdot|\mathbf{x})\), where $N$ is the per-position candidate budget.
Given a set of adversarial candidates, the attacker iteratively replaces the target tokens in \(S(\mathbf x)\) with each candidate to form a new potential adversarial document. This procedure is greedy in nature, and we retain the candidates that tend to decrease the true class probability the most. The procedure continues iterating over the target tokens until the prediction has changed or the maximum allowed number of queries is reached and the attack fails.

\section{Experiment}
\label{sec:exp}
We conduct experiments on both natural language inference and generative Q\&A tasks to demonstrate the versatility of our attacker. Following prior work \cite{li2020bertattack,liu2023sspattack,liu2023hqa}, we adopt the following metrics for fair comparison:

\noindent\textbf{Post-attack Accuracy.} For the classification attack, it is the fraction of adversarial documents still assigned their correct label. For the generative attack, it is the fraction of responses still judged safe by the guard model. A lower value therefore indicates a more effective attack.

\noindent\textbf{Number of Queries.} The average number of queries issued to the target model during an attack to create a single adversarial example. This definition is aligned with the baselines we are comparing to. We will show in the following section that we train our attacker against off-line and cheap target models, and therefore training queries are not a heavy concern.

\noindent\textbf{USE.} The cosine similarity between the original and perturbed texts, each embedded with the Universal Sentence Encoder~\cite{cer2018universal}. It measures how well the perturbation preserves the original meaning, so a higher value denotes better semantic preservation.

\noindent\textbf{Percentage of Perturbation.} The proportion of tokens altered relative to the length of the original document. A lower percentage reflects a smaller, less conspicuous edit and hence less semantic distortion.

\subsection{Implementation Details}
\label{sec:impl}
The same RL-ATTACK implementation drives both natural language inference (NLI) and generative Q\&A: the attacker architecture, the policy-gradient training objective (Eq.~\ref{eq:loglikelihood}), and the identify-influential-tokens-then-substitute attack mechanism (Section~\ref{subsec:attacking}) are identical across the two tasks---the framework is universal. The two tasks differ in two respects: the adversarial reward and the criterion for a successful attack. 

For NLI the target is a classifier: the reward is the drop in its true-label confidence $C(y_{\mathrm{true}}\mid\cdot)$, and an attack succeeds when the prediction flips to any incorrect label. 

For generative Q\&A, the ultimate attack target is an LLM whose responses are assessed by a safety judge. To reduce training costs, we do not query the target LLM during attacker training. Instead, we train the attacker against Llama-Guard-4-12B, a lower-cost safety classifier designed to identify unsafe prompts. We hypothesize that optimizing the attacker against this specialized judge enables it to learn adversarial perturbations that transfer to jailbreak attacks on target LLMs, and we evaluate this transfer at attack time. Specifically, during training, Llama-Guard-4-12B provides the confidence $C(y_{\mathrm{true}}\mid\cdot)$, defined as the probability assigned to a prompt's ground-truth safety label, which is used to compute the adversarial reward. At attack time, the resulting adversarial prompt is submitted to the target LLM, and the generated response is evaluated by either Llama-Guard-4-12B or the OpenAI Moderation API. A jailbreak is considered successful when the active judge classifies the target LLM's response as unsafe. This design still follows our generalized framework.

The attacker is a masked-language-model perturbation network built on BERT with its built-in masked-LM head. Of its $109.5$\,M total parameters, only ${\approx}0.6$\,M are trainable: the pretrained encoder and the tied token embeddings are frozen, and we learn only the masked-LM projection head (a $768\!\rightarrow\!768$ transform with an output bias over the $30{,}522$-token vocabulary). Given a document with $K$ masked positions, the head emits a distribution over the vocabulary at each position, from which the attack draws the top-$N$ candidate substitutions per position at inference. The maximum sequence length is $512$ tokens.

% \begin{table}[t]
%     \footnotesize
%     \caption{Natural-language-inference data splits.}
%     \label{tab:nli_data_splits}%
%     \centering
%     \setlength{\tabcolsep}{2pt}
%     \begin{tabular}{@{}>{\raggedright\arraybackslash}p{0.18\columnwidth} >{\raggedright\arraybackslash}p{0.25\columnwidth} >{\raggedright\arraybackslash}p{0.51\columnwidth}@{}}
%     \toprule
%     Dataset & Split / track & Purpose \\
%     \midrule
%     SNLI & \texttt{train} & Train the attacker against the SNLI BERT classifier \\
%     SNLI & \texttt{dev} & Select attack configuration and checkpoint \\
%     SNLI & held-out evaluation & Report SNLI attack results \\
%     MNLI & \texttt{train} & Train the attacker against the MNLI BERT classifier \\
%     MNLI matched & held-out evaluation & Report in-domain matched-genre attack results \\
%     MNLI mismatched & held-out evaluation & Report cross-genre mismatched attack results \\
%     \botrule
%     \end{tabular}
%     \end{table}

\begin{table*}[t]
\footnotesize
\caption{Adversarial attack results on \textbf{three Natural Language Inference datasets}. Post-attack accuracy is the percentage of adversarial documents still assigned their correct labels. Parenthetical values are standard errors computed from $10$ validation folds.}
\label{tab:nli_results}%
\centering
\setlength{\tabcolsep}{3pt}
\begin{tabular*}{\textwidth}{@{\extracolsep{\fill}}lllllll@{}}
\toprule
Dataset & \makecell{Target\\Model} & Method & \makecell{Post-attack\\accuracy (\%)} & \makecell{\# of\\queries} & USE & \makecell{\% of\\perturbation} \\
\midrule
SNLI & BERT & \textbf{RL-ATTACK} & 2.03~($\pm$0.54) & \textbf{13} & \textbf{0.65} & \textbf{5.1} \\
& & BERT (untrained) & 5.73 & 18 & 0.60 & 6.0 \\
               & & TextFooler \cite{jin2020bert} & 4.00 & 60 & 0.45 & 18.5 \\
               & & BERT-ATTACK \cite{li2020bertattack} & 5.73 & 16 & 0.40 & 12.4 \\
               & & SSPAttack \textbf{(hard-label)} \cite{liu2023sspattack} & \textbf{1.30} & 1000 & 0.54 & 15.9 \\
               & & HQA-Attack \textbf{(hard-label)} \cite{liu2023hqa} & \textbf{1.30} & 1000 & 0.54 & 15.9 \\
               & & TextHoaxer \textbf{(hard-label)} \cite{ye2022texthoaxer} & \textbf{1.30} & 1000 & 0.39 & 16.6 \\
               & & LeapAttack \textbf{(hard-label)} \cite{ye2022leapattack} & \textbf{1.30} & 1000 & 0.35 & 19.9 \\
\midrule
MNLI matched & BERT & \textbf{RL-ATTACK} & \textbf{0.90}~($\pm$0.30) & 15 & \textbf{0.70} & \textbf{3.2} \\
& & BERT (untrained) & 3.00 & \textbf{14} & 0.58 & 4.5 \\
               & & TextFooler & 9.60 & 78 & 0.57 & 15.2 \\
               & & BERT-ATTACK & 7.90 & 19 & 0.55 & 8.8 \\
               & & SSPAttack \textbf{(hard-label)} & 2.90 & 1000 & 0.64 & 12.0 \\
               & & HQA-Attack \textbf{(hard-label)} & 2.90 & 1000 & 0.64 & 12.1 \\
               & & TextHoaxer \textbf{(hard-label)} & 2.90 & 1000 & 0.53 & 12.7 \\
               & & LeapAttack \textbf{(hard-label)} & 2.90 & 1000 & 0.49 & 15.7 \\
\midrule
MNLI mismatched & BERT & \textbf{RL-ATTACK} & \textbf{0.80}~($\pm$0.28) & \textbf{16} & \textbf{0.73} & \textbf{3.0} \\
& & BERT (untrained) & 2.10 & 32 & 0.63 & 4.2 \\
               & & TextFooler & 8.3 & 86 & 0.58 & 14.6 \\
               & & BERT-ATTACK & 7.00 & 24 & 0.53 & 4.6 \\
               & & SSPAttack \textbf{(hard-label)} & 1.70 & 1000 & 0.65 & 11.5 \\
               & & HQA-Attack \textbf{(hard-label)} & 1.70 & 1000 & 0.65 & 11.5 \\
               & & TextHoaxer \textbf{(hard-label)} & 1.70 & 1000 & 0.54 & 12.5 \\
               & & LeapAttack \textbf{(hard-label)} & 1.70 & 1000 & 0.50 & 15.1 \\
\botrule
\end{tabular*}
\end{table*}

\subsection{Evaluation on Natural Language Inference}
\label{sec:exp_nli}
\noindent\textbf{Datasets.} We conduct untargeted adversarial attacks aiming to cause the classifier to predict any incorrect label on the standard SNLI \cite{bowman2015large} and MNLI \cite{williams-etal-2018-broad} benchmarks, with MNLI consisting of matched and mismatched tracks. These datasets have pre-split train, validation, and evaluation subsets.

For each NLI benchmark, the train split is used to train the attacker against the corresponding BERT classifier, the validation split is used to choose a checkpoint during training, and the evaluation set is used to report performance.

\noindent\textbf{Training setup.} During training, we allow the attack to modify tokens in both the premise and the hypothesis. This encourages the model to consider the full context of the document and learn more comprehensive adversarial strategies. However, during evaluation, we follow prior work \cite{li2020bertattack,liu2023sspattack,liu2023hqa} and attack the hypothesis only for the purpose of fair comparison.

\begin{table*}[t]
    \footnotesize
    \caption{Adversarial attack results on the \textbf{AdvBench dataset}. Post-attack accuracy is the percentage of responses still judged safe by the judge. Parenthetical values are standard errors computed from $10$ validation folds.}
    \label{tab:genai_results}%
    \centering
    \setlength{\tabcolsep}{4pt}
    \begin{tabular*}{\textwidth}{@{\extracolsep{\fill}}lllllll@{}}
    \toprule
    Judge & Target Model & Method & \makecell{Post-attack\\accuracy} & \makecell{\# of\\queries} & USE & \makecell{\% of\\perturbation}\\
    \midrule
    Llama Guard 4 12B & Llama 2 7B\cite{touvron2023llama} & \textbf{RL-ATTACK}  & \textbf{82.0}~($\pm$3.8) & \textbf{27.7} &  \textbf{0.98} & 2.0\% \\
    & & BERT (untrained) & 87.0 & 28.3 &  \textbf{0.98} & \textbf{1.0\%} \\
    & & PAIR\cite{chao2025jailbreaking}  & 98.0 & 56.0  & 0.60 & \text{N/A} \\
    \cmidrule(lr){2-7}
      & GPT-3.5-Turbo& \textbf{RL-ATTACK} &  \textbf{11.5}~($\pm$3.3) &  16.8 &  \textbf{0.98} &  \textbf{4.5\%} \\
     & & BERT (untrained) & 15.5 & \textbf{15.5} & 0.97 & 4.7\% \\
        & & PAIR & 28.0 & 29.6 & 0.44 & \text{N/A} \\
     \cmidrule(lr){2-7}
     & GPT-5.4-mini & \textbf{RL-ATTACK}& 92.6~($\pm$2.6)  & \textbf{29.0} & \textbf{0.99} & \textbf{1.1\%}  \\
     & & BERT (untrained) &  97.7 & 29.5  & 0.98 & 2.1\% \\
     & & PAIR & \textbf{77.6} & 52.6 &  0.48 & \text{N/A} \\
    \midrule
    OpenAI Moderation API& Llama 2 7B\cite{touvron2023llama} & \textbf{RL-ATTACK}  & \textbf{81.0}~($\pm$3.7) & \textbf{26.9} & \textbf{0.99} & \textbf{1.4\%} \\
    & & BERT (untrained) & 85.3 & \textbf{26.9} & 0.97 & 1.6\% \\
    & & PAIR & 88.0 & 29.3 & 0.53 & \text{N/A} \\
    \cmidrule(lr){2-7}
      & GPT-3.5-Turbo& \textbf{RL-ATTACK} & 42.0~($\pm$4.9) & \textbf{33.9}  & \textbf{0.96} & \textbf{4.1\%} \\
     & & BERT (untrained) & 61.5  & 47.0  &  \textbf{0.96} & 4.4\%  \\
        & & PAIR & \textbf{40.0}  & 46.6 & 0.46 & \text{N/A} \\
    \cmidrule(lr){2-7}
     & GPT-5.4-mini & \textbf{RL-ATTACK}& \textbf{84.0}~($\pm$3.7)  & \textbf{26.2} & \textbf{0.99} & 1.1\%  \\
     & & BERT (untrained) & 84.4 & 28.0  & 0.98  & \textbf{0.7\%} \\
     & & PAIR & 97.0  & 56.8 & 0.46  & \text{N/A} \\
    \botrule
    \end{tabular*}
    \end{table*}
    
\noindent\textbf{Training and attack parameters.} Across many training hyperparameter settings, our attacks efficiently reduce post-attack accuracy. We do not observe notably stronger performance when tuning the training hyperparameters. We therefore choose the final configuration of $\alpha=1$ and $K=3$ across all NLI tasks. This small edit budget helps achieve better semantic preservation while remaining sufficient to achieve low post-attack accuracy. For the attack hyperparameter, we set $N=40$ for all NLI tasks. We observe that this relatively small per-position candidate budget already reduces post-attack accuracy to approximately $2\%$ or lower as shown in Table~\ref{tab:nli_results}. We therefore choose this small value to use fewer queries and better preserve semantics.

We will discuss the hyperparameters for the generative Q\&A task in greater detail in the following section, since we observed more obvious performance changes when tuning parameters for it.

\noindent\textbf{Main results.} As detailed in Table~\ref{tab:nli_results}, RL-ATTACK consistently achieves the lowest post-attack accuracies among the soft-label attacks, with all three values around $2\%$ or below, outperforming the soft-label baselines. Relative to the BERT (untrained) baseline, RL-ATTACK lowers post-attack accuracy on all three NLI evaluation tracks. We hypothesize that these gains result from training the attacker from a vanilla BERT initialization.

We also include hard-label attacks to position our results against the broader adversarial-attack literature, but they should not be read as a fair like-for-like comparison: hard-label methods face a different feedback setting because they do not access predicted-label probabilities, and the reported hard-label baselines use much larger query budgets. On SNLI, several hard-label methods attain lower raw post-attack accuracy than RL-ATTACK, but only with $1000$ queries and lower semantic similarity; on MNLI matched and mismatched, RL-ATTACK has the lowest raw post-attack accuracy while also using far fewer queries. While the semantic similarity---measured by the USE score---varies across datasets, RL-ATTACK keeps the query count and perturbation rate low given the strength of the attack. Across all datasets, it requires, on average, only 13 to 16 queries to successfully attack a hypothesis, with perturbation rates largely remaining around 5\% or below. 

\subsection{Evaluation on Generative Q\&A}
\label{sec:exp_genqa}
\noindent\textbf{Datasets.} We use the in-the-wild jailbreak-prompt corpus \cite{SCBSZ24} and AdvBench \cite{zou2023universal}. We partition the in-the-wild dataset into three disjoint splits: \texttt{train}, \texttt{traindev}, \texttt{evaldev}. The \texttt{train} split is used to train the attacker, \texttt{traindev} is used for checkpoint selection, and \texttt{evaldev} is used to select the attacking hyperparameters. The same trained checkpoint is evaluated on AdvBench since AdvBench itself is a small dataset that is built specifically for robustness evaluation purposes. Table~\ref{tab:data_splits} summarizes the splits.

\begin{table}[t]
\footnotesize
\caption{Generative Q\&A attack data splits.}
\label{tab:data_splits}%
\centering
\setlength{\tabcolsep}{2pt}
\begin{tabular}{@{}>{\raggedright\arraybackslash}p{0.21\columnwidth} >{\raggedright\arraybackslash}p{0.17\columnwidth} >{\raggedleft\arraybackslash}p{0.13\columnwidth} >{\raggedright\arraybackslash}p{0.42\columnwidth}@{}}
\toprule
Dataset & Split & Size & Purpose \\
\midrule
in-the-wild & \texttt{train} & $16{,}776$ & Train the attacker \\
in-the-wild & \texttt{traindev} & $50$ & Checkpoint selection\\
in-the-wild & \texttt{evaldev} & $100$ & Attack hyperparameter selection\\
AdvBench & \texttt{all} & $500$ &  Final result reported \\
\botrule
\end{tabular}
\end{table}

\noindent\textbf{Target models.} We evaluate the vulnerability of Llama 2 7B~\cite{touvron2023llama}, GPT-3.5-Turbo, and GPT-5.4-mini. These models are selected to ensure a fair comparison with PAIR~\cite{chao2025jailbreaking} and to assess transferability across model scales. We do not evaluate more advanced proprietary models (e.g., GPT-5, Claude, Gemini~Pro): their providers deploy aggressive, automated account-banning mechanisms, so issuing even a handful of adversarial (policy-violating) prompts risks suspending or banning the account, which makes systematic and repeatable evaluation infeasible.

\noindent\textbf{Reward and judge models.} Section~\ref{sec:impl} summarizes the reward and success criterion for generative targets. Here, we specify how the judge confidence instantiates the shared objective. During training, Llama-Guard-4-12B~\cite{inan2023llamaguard} provides the true-label confidence $C(y_{\mathrm{true}}\mid\cdot)$. The three reward variants (Table~\ref{tab:reward_variants}), the semantic term $r_{\mathrm{sem}}$ (USE cosine similarity), and the combined reward $r=\alpha r_{\mathrm{adv}}+(1-\alpha)r_{\mathrm{sem}}$ are exactly as defined in Section~\ref{sec:training_objective}, with $\alpha\in\{0.0,0.1,\dots,1.0\}$ sweeping the trade-off between attack effectiveness and semantic similarity. 

At attack time, we first evaluate RL-ATTACK using Llama-Guard-4-12B, the same judge used during training and the judge adopted by PAIR. We additionally use the OpenAI Moderation API as a supplementary judge to test whether the attacker trained with the Llama Guard reward generalizes when attack success is determined by a different safety model. The Moderation API is never used during training. 

\begin{figure}[t]
    \centering
    \includegraphics[width=\linewidth]{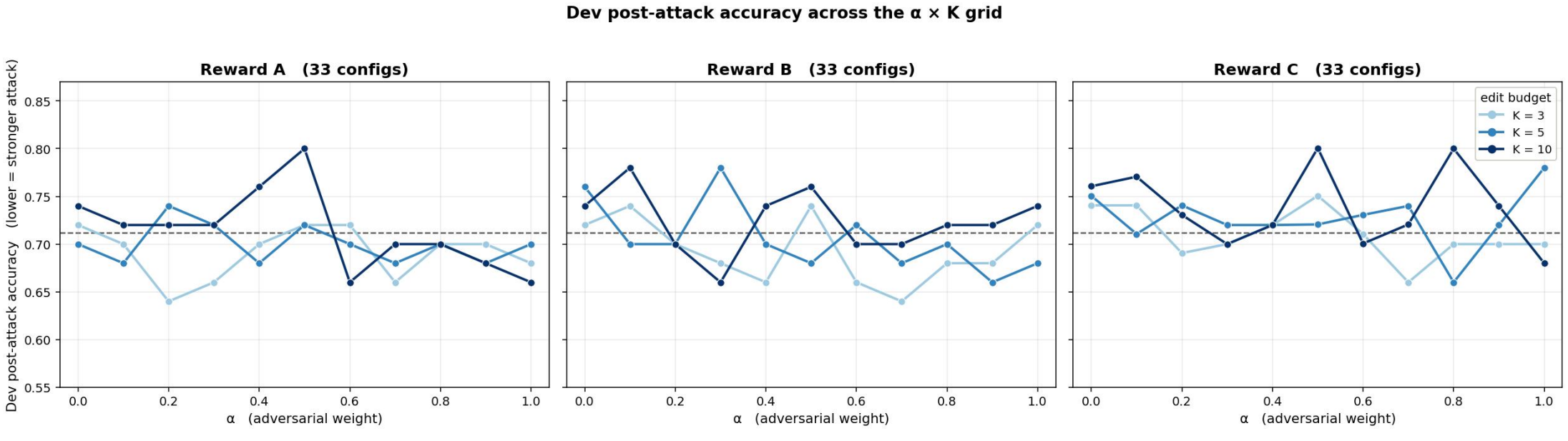}
    \caption{Post-attack accuracy on the in-the-wild \texttt{traindev} set. Each point represents a combination of the reward variant, $\alpha$, and $K$. No consistent trend is observed.}
    \label{fig:grid}
\end{figure}

\noindent\textbf{Training and attacking.} We optimize only the projection head using Adam (learning rate $0.01$, $\beta=(0.9,0.98)$) with batch size $b=16$ for one epoch ($\approx\!1{,}049$ steps over the $16{,}776$ training prompts). We check the held-out DEV reward (the same objective on the $50$-prompt \texttt{traindev} split) every $100$ steps and save the checkpoint whenever it improves. The training hyperparameters are chosen by grid search over $\mathrm{reward}\in\{A,B,C\}\times\alpha\in\{0.0,\dots,1.0\}\times K\in\{3,5,10\}$ ($99$ configurations, with the candidate budget fixed at $N=1$ during the search), each scored by $10$-fold cross-validation on \texttt{traindev} (Figure~\ref{fig:grid}); the edit budget $K$ ($=$\,\texttt{num\_doc\_masks}) is a training hyperparameter, whereas the candidate budget $N$ ($=$\,\texttt{samples\_per\_tok}) is fixed only at attack time. At attack time, we run a greedy token-substitution search with edit budget $K$ and search budget $N$; a document is successfully attacked when the target LLM's response (temperature $0$, \texttt{max\_new\_tokens}$=96$) is judged unsafe by the active judge, and we report accuracy after attack.

\begin{figure}[t]
    \centering
    \includegraphics[width=\linewidth]{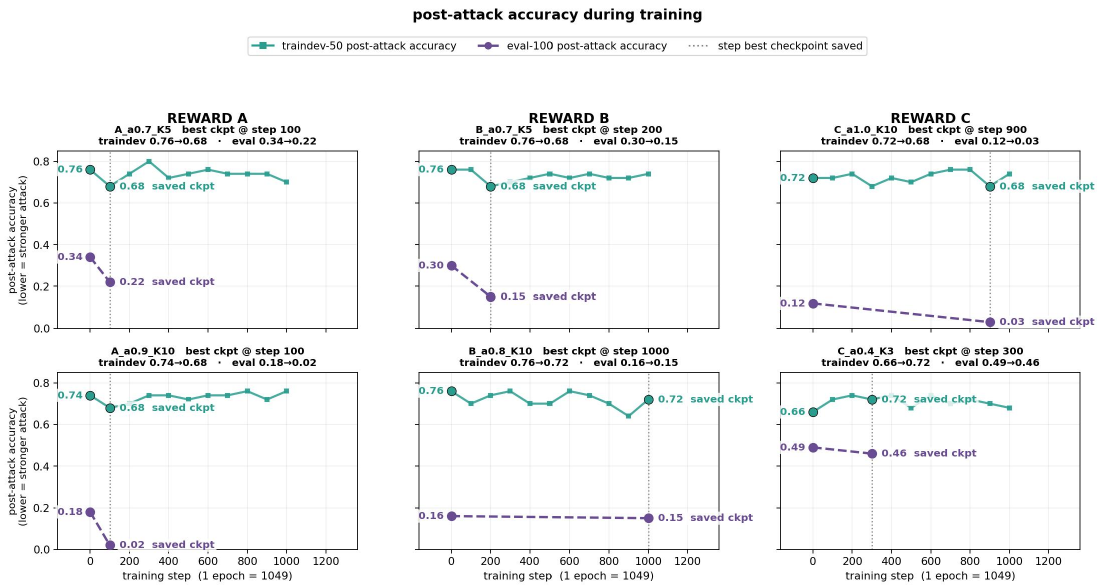}
    \caption{Post-attack accuracy over training for six representative hyperparameter combinations under three types of rewards.}
    \label{fig:train_curves}
\end{figure}

\noindent\textbf{Hyperparameter sensitivity and training effect.} Fig.~\ref{fig:grid} summarizes $99$ configurations spanning three reward variants, $\alpha\in\{0.0,0.1,\dots,1.0\}$, and $K\in\{3,5,10\}$. Their \texttt{traindev} post-attack accuracies range from $0.64$ to $0.80$ (mean $0.71$, SD $0.035$), with no consistent advantage for any reward variant, value of $\alpha$, or edit budget. We therefore treat the grid as a descriptive sensitivity analysis, not evidence of a unique optimum.

Fig.~\ref{fig:train_curves} compares the untrained attacker with the selected checkpoint for six representative configurations. Evaluation post-attack accuracy decreases in all six, from a marginal $0.16\rightarrow0.15$ to a substantial $0.18\rightarrow0.02$, while \texttt{traindev} accuracy decreases in five. In the remaining case, \texttt{traindev} accuracy increases from $0.66$ to $0.72$ even as evaluation accuracy decreases from $0.49$ to $0.46$. Moreover, four checkpoints with the same \texttt{traindev} accuracy of $0.68$ yield evaluation accuracies from $0.02$ to $0.22$. Thus, training strengthens the attack in these examples, but \texttt{traindev} accuracy does not reliably rank held-out performance. 

We therefore use reward variant~A with $\alpha=1$ and $K=3$ as a simple, competitive attacker-training configuration with the default reward and smallest tested edit budget, without claiming a unique optimum.

\noindent\textbf{Attack-time candidate budget ($N$).} During attacker training and training-hyperparameter evaluation, we hold $N$ at $1$. At attack time, we observe diminishing returns as $N$ increases. We set $N=40$ in all generative LLM attack settings because it provides a good trade-off between post-attack accuracy and query cost.

\noindent\textbf{Main results.} Table~\ref{tab:genai_results} reports attacks on three target models under two independent judges. Only Llama Guard 4 12B is used for training the attacker. The OpenAI Moderation API is used to demonstrate the generalization of our attacker. Compared with the optimization-based PAIR baseline, RL-ATTACK uses fewer queries and achieves substantially higher semantic similarity in all six settings: its USE scores range from $0.96$ to $0.99$, compared with $0.44$ to $0.60$ for PAIR. RL-ATTACK also achieves lower post-attack accuracy than PAIR in four of the six settings. The two exceptions are GPT-5.4-mini under Llama Guard ($92.6\%$ for RL-ATTACK versus $77.6\%$ for PAIR) and GPT-3.5-Turbo under the Moderation API ($42.0\%$ versus $40.0\%$). 

Relative to the untrained BERT attacker, RL-ATTACK achieves a lower post-attack-accuracy point estimate in all six settings while maintaining comparable USE scores and query efficiency. These results support the conclusion that reinforcement-learning training generally improves attack effectiveness without sacrificing semantic preservation or query efficiency, although the improvement over the untrained attacker is small for GPT-5.4-mini under the Moderation API ($84.0\%$ versus $84.4\%$). GPT-5.4-mini remains the most difficult target overall in our setting.

\section{Conclusion}
We presented RL-ATTACK, a trainable black-box adversarial framework that leverages reinforcement learning to learn target-model-specific transformations. Experiments on natural language inference and generative Q\&A tasks show that RL-ATTACK produces strong attacks with high semantic preservation at low query cost in most evaluation settings and improves over an untrained attacker. Our conclusions are limited by the scope of our evaluation settings, the modest sizes of the evaluation sets, and the lack of an explicit grammaticality constraint on adversarial documents.
\section*{Acknowledgment}
This work was supported by the Financial Services Analytics (FSAN) Ph.D. Program at the University of Delaware.
\balance
\bibliographystyle{IEEEtran}
\bibliography{mybibliography}

@inproceedings{hu2018blackbox,
    title = {Black-box attacks against RNN based malware detection algorithms},
    author = {Hu, W. and Tan, Y.},
    booktitle = {Workshops at the Thirty-Second AAAI Conference on Artificial Intelligence},
    year = {2018},
    month = jun
}

@inproceedings{gao2018blackbox,
    title = {Black-box generation of adversarial text sequences to evade deep learning classifiers},
    author = {Gao, J. and Lanchantin, J. and Soffa, M. L. and Qi, Y.},
    booktitle = {2018 IEEE Security and Privacy Workshops (SPW)},
    pages = {50--56},
    year = {2018},
    month = may,
    publisher = {IEEE}
}

@article{li2018textbugger,
    title = {Textbugger: Generating adversarial text against real-world applications},
    author = {Li, J. and Ji, S. and Du, T. and Li, B. and Wang, T.},
    year = {2018},
    journal = {arXiv preprint arXiv:1812.05271}
}

@misc{li2020bertattack,
    title={Bert-attack: Adversarial attack against bert using bert},
    author={Li, Linyang and Ma, Ruotian and Guo, Qipeng and Xue, Xiang and Qiu, Xipeng},
    year={2020},
    eprint={2004.09984},
    archivePrefix={arXiv},
    primaryClass={cs.CL}
}

@inproceedings{cer2018universal,
    title = "Universal Sentence Encoder for {E}nglish",
    author = "Cer, Daniel  and
      Yang, Yinfei  and
      Kong, Sheng-yi  and
      Hua, Nan  and
      Limtiaco, Nicole  and
      St. John, Rhomni  and
      Constant, Noah  and
      Guajardo-Cespedes, Mario  and
      Yuan, Steve  and
      Tar, Chris  and
      Strope, Brian  and
      Kurzweil, Ray",
    editor = "Blanco, Eduardo  and
      Lu, Wei",
    booktitle = "Proceedings of the 2018 Conference on Empirical Methods in Natural Language Processing: System Demonstrations",
    month = nov,
    year = "2018",
    address = "Brussels, Belgium",
    publisher = "Association for Computational Linguistics",
    url = "https://aclanthology.org/D18-2029/",
    doi = "10.18653/v1/D18-2029",
    pages = "169--174"
}

@inproceedings{jin2020bert,
  title={Is bert really robust? a strong baseline for natural language attack on text classification and entailment},
  author={Jin, Di and Jin, Zhijing and Zhou, Joey Tianyi and Szolovits, Peter},
  booktitle={Proceedings of the AAAI conference on artificial intelligence},
  volume={34},
  pages={8018--8025},
  year={2020}
}

@article{koroteev2021bert,
  title={BERT: a review of applications in natural language processing and understanding},
  author={Koroteev, Mikhail V},
  journal={arXiv preprint arXiv:2103.11943},
  year={2021}
}

@article{yao2024survey,
  title={A survey on large language model (llm) security and privacy: The good, the bad, and the ugly},
  author={Yao, Yifan and Duan, Jinhao and Xu, Kaidi and Cai, Yuanfang and Sun, Zhibo and Zhang, Yue},
  journal={High-Confidence Computing},
  pages={100211},
  year={2024},
  publisher={Elsevier}
}

@inproceedings{wu2018learning,
  title={Learning to extract coherent summary via deep reinforcement learning},
  author={Wu, Yuxiang and Hu, Baotian},
  booktitle={Proceedings of the AAAI conference on artificial intelligence},
  volume={32},
  year={2018}
}

@article{paulus2017deep,
  title={A deep reinforced model for abstractive summarization},
  author={Paulus, R},
  journal={arXiv preprint arXiv:1705.04304},
  year={2017}
}

@article{gao2020preference,
  title={Preference-based interactive multi-document summarisation},
  author={Gao, Yang and Meyer, Christian M and Gurevych, Iryna},
  journal={Information Retrieval Journal},
  volume={23},
  pages={555--585},
  year={2020},
  publisher={Springer}
}

@inproceedings{nguyen2017reinforcement,
  title     = {Reinforcement Learning for Bandit Neural Machine Translation with Simulated Human Feedback},
  author    = {Nguyen, Khanh and Daum{\'e} III, Hal and Boyd-Graber, Jordan},
  booktitle = {Proceedings of the 2017 Conference on Empirical Methods in Natural Language Processing (EMNLP)},
  year      = {2017},
  url       = {https://aclanthology.org/D17-1153/}
}

@article{ouyang2022training,
  title={Training language models to follow instructions with human feedback},
  author={Ouyang, Long and Wu, Jeffrey and Jiang, Xu and Almeida, Diogo and Wainwright, Carroll and Mishkin, Pamela and Zhang, Chong and Agarwal, Sandhini and Slama, Katarina and Ray, Alex and others},
  journal={Advances in neural information processing systems},
  volume={35},
  pages={27730--27744},
  year={2022}
}

@article{achiam2023gpt,
  title  = {{GPT-4} Technical Report},
  author = {{OpenAI}},
  journal = {arXiv preprint arXiv:2303.08774},
  year   = {2023},
  eprint = {2303.08774},
  archivePrefix = {arXiv},
  primaryClass = {cs.CL},
  url    = {https://arxiv.org/abs/2303.08774}
}

@article{touvron2023llama,
  title={Llama 2: Open foundation and fine-tuned chat models},
  author={Touvron, Hugo and Martin, Louis and Stone, Kevin and Albert, Peter and Almahairi, Amjad and Babaei, Yasmine and Bashlykov, Nikolay and Batra, Soumya and Bhargava, Prajjwal and Bhosale, Shruti and others},
  journal={arXiv preprint arXiv:2307.09288},
  year={2023}
}

@article{sutton1999policy,
  title={Policy gradient methods for reinforcement learning with function approximation},
  author={Sutton, Richard S and McAllester, David and Singh, Satinder and Mansour, Yishay},
  journal={Advances in neural information processing systems},
  volume={12},
  year={1999}
}

@inproceedings{ye2022texthoaxer,
  title={Texthoaxer: Budgeted hard-label adversarial attacks on text},
  author={Ye, Muchao and Miao, Chenglin and Wang, Ting and Ma, Fenglong},
  booktitle={Proceedings of the AAAI Conference on Artificial Intelligence},
  volume={36},
  pages={3877--3884},
  year={2022}
}

@article{team2023gemini,
  title={Gemini: a family of highly capable multimodal models},
  author={Team, Gemini and Anil, Rohan and Borgeaud, Sebastian and Alayrac, Jean-Baptiste and Yu, Jiahui and Soricut, Radu and Schalkwyk, Johan and Dai, Andrew M and Hauth, Anja and Millican, Katie and others},
  journal={arXiv preprint arXiv:2312.11805},
  year={2023}
}

@inproceedings{liu2023sspattack,
  title={SSPAttack: a simple and sweet paradigm for black-box hard-label textual adversarial attack},
  author={Liu, Han and Xu, Zhi and Zhang, Xiaotong and Xu, Xiaoming and Zhang, Feng and Ma, Fenglong and Chen, Hongyang and Yu, Hong and Zhang, Xianchao},
  booktitle={Proceedings of the AAAI conference on artificial intelligence},
  volume={37},
  pages={13228--13235},
  year={2023}
}

@article{liu2023hqa,
  title={HQA-attack: toward high quality black-box hard-label adversarial attack on text},
  author={Liu, Han and Xu, Zhi and Zhang, Xiaotong and Zhang, Feng and Ma, Fenglong and Chen, Hongyang and Yu, Hong and Zhang, Xianchao},
  journal={Advances in Neural Information Processing Systems},
  volume={36},
  pages={51347--51358},
  year={2023}
}

@inproceedings{SCBSZ24,
      author = {Xinyue Shen and Zeyuan Chen and Michael Backes and Yun Shen and Yang Zhang},
      title = {{``Do Anything Now'': Characterizing and Evaluating In-The-Wild Jailbreak Prompts on Large Language Models}},
      booktitle = {{ACM SIGSAC Conference on Computer and Communications Security (CCS)}},
      publisher = {ACM},
      year = {2024}
}

@misc{zou2023universal,
      title={Universal and Transferable Adversarial Attacks on Aligned Language Models}, 
      author={Andy Zou and Zifan Wang and J. Zico Kolter and Matt Fredrikson},
      year={2023},
      eprint={2307.15043},
      archivePrefix={arXiv},
      primaryClass={cs.CL}
}

@inproceedings{alzantot-etal-2018-generating,
    title = "Generating Natural Language Adversarial Examples",
    author = "Alzantot, Moustafa  and
      Sharma, Yash  and
      Elgohary, Ahmed  and
      Ho, Bo-Jhang  and
      Srivastava, Mani  and
      Chang, Kai-Wei",
    editor = "Riloff, Ellen  and
      Chiang, David  and
      Hockenmaier, Julia  and
      Tsujii, Jun{'}ichi",
    booktitle = "Proceedings of the 2018 Conference on Empirical Methods in Natural Language Processing",
    month = oct # "-" # nov,
    year = "2018",
    address = "Brussels, Belgium",
    publisher = "Association for Computational Linguistics",
    url = "https://aclanthology.org/D18-1316/",
    doi = "10.18653/v1/D18-1316",
    pages = "2890--2896"
}

@article{grattafiori2024llama,
  title={The llama 3 herd of models},
  author={Grattafiori, Aaron and Dubey, Abhimanyu and Jauhri, Abhinav and Pandey, Abhinav and Kadian, Abhishek and Al-Dahle, Ahmad and Letman, Aiesha and Mathur, Akhil and Schelten, Alan and Vaughan, Alex and others},
  journal={arXiv preprint arXiv:2407.21783},
  year={2024}
}

@inproceedings{li2020contextualized,
    title = "Contextualized Perturbation for Textual Adversarial Attack",
    author = "Li, Dianqi  and
      Zhang, Yizhe  and
      Peng, Hao  and
      Chen, Liqun  and
      Brockett, Chris  and
      Sun, Ming-Ting  and
      Dolan, Bill",
    editor = "Toutanova, Kristina  and
      Rumshisky, Anna  and
      Zettlemoyer, Luke  and
      Hakkani-Tur, Dilek  and
      Beltagy, Iz  and
      Bethard, Steven  and
      Cotterell, Ryan  and
      Chakraborty, Tanmoy  and
      Zhou, Yichao",
    booktitle = "Proceedings of the 2021 Conference of the North American Chapter of the Association for Computational Linguistics: Human Language Technologies",
    month = jun,
    year = "2021",
    address = "Online",
    publisher = "Association for Computational Linguistics",
    url = "https://aclanthology.org/2021.naacl-main.400/",
    doi = "10.18653/v1/2021.naacl-main.400",
    pages = "5053--5069"
}

@inproceedings{reimers2019sentence,
    title = "Sentence-{BERT}: Sentence Embeddings using {S}iamese {BERT}-Networks",
    author = "Reimers, Nils  and
      Gurevych, Iryna",
    editor = "Inui, Kentaro  and
      Jiang, Jing  and
      Ng, Vincent  and
      Wan, Xiaojun",
    booktitle = "Proceedings of the 2019 Conference on Empirical Methods in Natural Language Processing and the 9th International Joint Conference on Natural Language Processing (EMNLP-IJCNLP)",
    month = nov,
    year = "2019",
    address = "Hong Kong, China",
    publisher = "Association for Computational Linguistics",
    url = "https://aclanthology.org/D19-1410/",
    doi = "10.18653/v1/D19-1410",
    pages = "3982--3992"
}

@article{liu2019roberta,
    title  = {{RoBERTa}: A Robustly Optimized {BERT} Pretraining Approach},
    author = {Liu, Yinhan and Ott, Myle and Goyal, Naman and Du, Jingfei and Joshi, Mandar and Chen, Danqi and Levy, Omer and Lewis, Mike and Zettlemoyer, Luke and Stoyanov, Veselin},
    journal = {arXiv preprint arXiv:1907.11692},
    year   = {2019},
    url    = {https://arxiv.org/abs/1907.11692}
}

@article{keneshloo2018deep,
 title={Deep Reinforcement Learning For Sequence to Sequence Models},
 author={Keneshloo, Yaser and Shi, Tian and Ramakrishnan, Naren and Reddy, Chandan K.},
 journal={arXiv preprint arXiv:1805.09461},
 year={2018}
}

@inproceedings{bowman2015large,
  title={A large annotated corpus for learning natural language inference},
  author={Bowman, Samuel and Angeli, Gabor and Potts, Christopher and Manning, Christopher D},
  booktitle={Proceedings of the 2015 conference on empirical methods in natural language processing},
  pages={632--642},
  year={2015}
}

@inproceedings{williams-etal-2018-broad,
    title = "A Broad-Coverage Challenge Corpus for Sentence Understanding through Inference",
    author = "Williams, Adina  and
      Nangia, Nikita  and
      Bowman, Samuel",
    editor = "Walker, Marilyn  and
      Ji, Heng  and
      Stent, Amanda",
    booktitle = "Proceedings of the 2018 Conference of the North {A}merican Chapter of the Association for Computational Linguistics: Human Language Technologies, Volume 1 (Long Papers)",
    month = jun,
    year = "2018",
    address = "New Orleans, Louisiana",
    publisher = "Association for Computational Linguistics",
    url = "https://aclanthology.org/N18-1101/",
    doi = "10.18653/v1/N18-1101",
    pages = "1112--1122"
}

@inproceedings{devlin2019bert,
  title={Bert: Pre-training of deep bidirectional transformers for language understanding},
  author={Devlin, Jacob and Chang, Ming-Wei and Lee, Kenton and Toutanova, Kristina},
  booktitle={Proceedings of the 2019 conference of the North American chapter of the association for computational linguistics: human language technologies, volume 1 (long and short papers)},
  pages={4171--4186},
  year={2019}
}

@inproceedings{chao2025jailbreaking,
  title={Jailbreaking black box large language models in twenty queries},
  author={Chao, Patrick and Robey, Alexander and Dobriban, Edgar and Hassani, Hamed and Pappas, George J and Wong, Eric},
  booktitle={2025 IEEE Conference on Secure and Trustworthy Machine Learning (SaTML)},
  pages={23--42},
  year={2025},
  organization={IEEE}
}

@inproceedings{ye2022leapattack,
  title={LeapAttack: Hard-Label Adversarial Attack on Text via Gradient-Based Optimization},
  author={Ye, Muchao and Chen, Jinghui and Miao, Chenglin and Wang, Ting and Ma, Fenglong},
  booktitle={Proceedings of the 28th ACM SIGKDD Conference on Knowledge Discovery and Data Mining},
  pages={2307--2315},
  year={2022}
}

@article{inan2023llamaguard,
  title={Llama Guard: {LLM}-based Input-Output Safeguard for Human-{AI} Conversations},
  author={Inan, Hakan and others},
  journal={arXiv preprint arXiv:2312.06674},
  year={2023}
}

\end{document}